\documentclass[letterpaper, 10 pt, conference]{ieeeconf}  

\IEEEoverridecommandlockouts                              

\usepackage{graphicx} 
\usepackage{amsmath} 
\usepackage{amssymb}  
\usepackage[table]{xcolor}
\usepackage{booktabs}
\usepackage{tabularx} 
\usepackage{scalerel}
\usepackage{tikz, pgfplots}
  \usetikzlibrary{svg.path}
\usepackage{url}

\makeatletter
\let\NAT@parse\undefined
\makeatother
\usepackage{hyperref}

\newcommand\copyrighttext{%
    \footnotesize \copyright{ }2025 IEEE. Personal use of this material is permitted. Permission from IEEE must be obtained for all other uses, in any current or future media, including reprinting/republishing this material for advertising or promotional purposes, creating new collective works, for resale or redistribution to servers or lists, or reuse of any copyrighted component of this work in other works.}
\newcommand\copyrightnotice{%
    \begin{tikzpicture}[remember picture,overlay]
    \node[anchor=south,yshift=15pt,xshift=0pt] at (current page.south) {\parbox{\dimexpr\textwidth-\fboxsep-\fboxrule\relax}{\copyrighttext}};
    \end{tikzpicture}%
}

\definecolor{orcidlogocol}{HTML}{A6CE39}
\tikzset{
  orcidlogo/.pic={
    \fill[orcidlogocol] svg{M256,128c0,70.7-57.3,128-128,128C57.3,256,0,198.7,0,128C0,57.3,57.3,0,128,0C198.7,0,256,57.3,256,128z};
    \fill[white] svg{M86.3,186.2H70.9V79.1h15.4v48.4V186.2z}
                 svg{M108.9,79.1h41.6c39.6,0,57,28.3,57,53.6c0,27.5-21.5,53.6-56.8,53.6h-41.8V79.1z M124.3,172.4h24.5c34.9,0,42.9-26.5,42.9-39.7c0-21.5-13.7-39.7-43.7-39.7h-23.7V172.4z}
                 svg{M88.7,56.8c0,5.5-4.5,10.1-10.1,10.1c-5.6,0-10.1-4.6-10.1-10.1c0-5.6,4.5-10.1,10.1-10.1C84.2,46.7,88.7,51.3,88.7,56.8z};
  }
}
\newcommand\orcidicon[1]{\href{https://orcid.org/#1}{\mbox{\scalerel*{
\begin{tikzpicture}[yscale=-1,transform shape]
\pic{orcidlogo};
\end{tikzpicture}
}{|}}}}

\title{\LARGE \bf
Application Management in C-ITS:\\ Orchestrating Demand-Driven Deployments and Reconfigurations
}

\author{
    Lukas Zanger\textsuperscript{\orcidicon{0009-0005-0292-2660}}*,
    Bastian Lampe\textsuperscript{\orcidicon{0000-0002-4414-6947}}*,
    Lennart Reiher\textsuperscript{\orcidicon{0000-0002-7309-164X}}*,
    and Lutz Eckstein
    \thanks{*These authors contributed equally to this work. \newline All authors are with the Institute for Automotive Engineering~(ika), RWTH Aachen University, 52074 Aachen, Germany. {\tt\small \{firstname.lastname\}@ika.rwth-aachen.de}}%
}

\begin{document}

\bstctlcite{IEEEexample:BSTcontrol}
\maketitle
\thispagestyle{empty}
\pagestyle{empty}
\copyrightnotice

\begin{abstract}
Vehicles are becoming increasingly automated and interconnected, enabling the formation of cooperative intelligent transport systems (\mbox{C-ITS}) and the use of offboard services. As a result, cloud-native techniques, such as microservices and container orchestration, play an increasingly important role in their operation.
However, orchestrating applications in a large-scale \mbox{C-ITS} poses unique challenges due to the dynamic nature of the environment and the need for efficient resource utilization.
In this paper, we present a demand-driven application management approach that leverages cloud-native techniques~--~specifically Kubernetes~--~to address these challenges. Taking into account the demands originating from different entities within the \mbox{C-ITS}, the approach enables the automation of processes, such as deployment, reconfiguration, update, upgrade, and scaling of microservices.
Executing these processes on demand can, for example, reduce computing resource consumption and network traffic.
A demand may include a request for provisioning an external supporting service, such as a collective environment model.
The approach handles changing and new demands by dynamically reconciling them through our proposed application management framework built on Kubernetes and the Robot Operating System (ROS 2).
We demonstrate the operation of our framework in the \mbox{C-ITS} use case of collective environment perception and make the source code of the prototypical framework publicly available at \href{https://github.com/ika-rwth-aachen/application_manager}{\nolinkurl{https://github.com/ika-rwth-aachen/application_manager}}.
\end{abstract}

\section{Introduction}
In future cooperative intelligent transport systems (\mbox{C-ITS}), various entities, such as vehicles equipped with driving automation systems, sensor-equipped roadside infrastructure units, edge/cloud servers, and control centers, will be connected, exchange data, and may offer computational resources.
These advancements enable new applications~--~such as collective environment perception, cooperative decision-making, computation offloading, and intelligent traffic management~--~that can contribute to improved comfort and safety for road users~\cite{wang_homogeneous_2024}, \cite{liu_behavioral_2023}, \cite{kim_impact_2015}.
Not only cloud and edge servers but also vehicles and roadside units can be part of a distributed computing system.
However, these applications may also introduce complexity that is difficult to manage. The dynamic nature of \mbox{C-ITS}, the presence of resource-constrained entities, and the strict requirements for safety and security pose unique challenges.

Cloud-native techniques provide a promising foundation for the development and operation of scalable applications in dynamic environments. Such techniques involve paradigms like containerization, microservice architectures, and container orchestration. They enable loosely coupled systems which are manageable and resilient~\cite{cloud_native_computing_foundation}. Said techniques and paradigms have the potential to contribute to the advancement of \mbox{C-ITS}.

Kubernetes has evolved as the de facto standard for orchestrating containerized applications in distributed systems. It is open-source and widely adopted by software companies worldwide~\cite{burnsKubernetesRunningDive2019}.
Nevertheless, Kubernetes lacks methods that are domain-specific, e.g., to \mbox{C-ITS}, considering that specific tasks like the deployment of required applications are only needed at certain times or may depend on the specific content of data exchanged in the \mbox{C-ITS}.
We have developed the approach \textit{RobotKube}~\cite{robotkube} to extend the regular capabilities of Kubernetes. RobotKube comprises software components designed to automate the identification of requirements and the formulation of specific Kubernetes workloads.
These components include the \textit{event detector} and the \textit{application manager}.

In this paper, we propose a demand-driven application management approach and present the methodology behind the application manager as part of an application management framework.
This methodology integrates seamlessly into the RobotKube architecture and complements parts of RobotKube which were not detailed yet.
The application management framework~--~comprising the application manager and a set of \textit{custom operators}~--~addresses the orchestration challenges in \mbox{C-ITS} through a demand-driven approach.
In this context, applications are deployed, reconfigured, scaled, and updated based on the current demands of \mbox{C-ITS} entities.

With our work, we make the following main contributions:
\begin{itemize}
    \item Presentation of the methodology for demand-driven application management allowing to deploy, reconfigure, update, upgrade, and scale applications based on demands of entities in a \mbox{C-ITS}.
    \item Prototypical implementation of the application manager and the custom operators in an application management framework based on Kubernetes and ROS 2.
    \item Demonstration and evaluation of the capabilities of the application management framework in the complex \mbox{C-ITS} use case of collective environment perception involving various \mbox{C-ITS} entities.
    \item Open source code publication of the application management framework and the demonstration use case allowing for reproducibility and extensibility.
    \item Complementation of RobotKube~\cite{robotkube} by providing the concrete methodology of the application manager.
\end{itemize}

\section{Related Work and Fundamentals}
\label{sec:related_work}
This work extends and further details our general orchestration approach introduced in RobotKube~\cite{robotkube}.
A central element of this approach is the \textit{operator application}, formed by the application manager and the event detector with its operator plugin.
Cluster operations in Kubernetes including the deployment and configuration of applications can be automated through operator applications.
Operator applications act on events, which are occurrences of patterns in data, in the cluster. Events are detected by event detector components which can be implemented in the scope of a framework for event detection in \mbox{C-ITS} presented in~\cite{reiherevent}. As outlined in~\cite{robotkube}, the application manager receives a task description from the event detector's operator plugin which the application manager then translates into a Kubernetes workload definition.
The application manager composes the requested applications from available microservices and deploys them to nodes in the Kubernetes cluster. It is also capable of managing existing applications it has launched. The actual methodology how the application manager works is not yet described in~\cite{robotkube}. The detailed methodology is one of the contributions of this paper.

Our application management approach relies on concepts like containerization, microservices, and container orchestration. Such concepts are defined by the Cloud Native Computing Foundation (CNCF) as cloud-native techniques~\cite{cloud_native_computing_foundation}. According to~\cite{cncf_container_orchestration}, \textit{containerization} ``is the process of packaging applications including their code and dependencies into single lightweight executables called container images''.
Software platforms exist to run the process of containerization and the deployment of containers~\cite{matthias_docker_2016}.
In \textit{microservices} architectures, applications are built as a set of loosely coupled, ``individual, independent (micro)services, with each service focused on a specific functionality''~\cite{cncf_microservices_architecture}. Separating functionality into distinct services makes them easier to deploy, update, and scale independently.

\textit{Orchestration} in the context of cloud-native techniques ``refers to managing and automating the lifecycle of containerized applications in dynamic environments''~\cite{cncf_container_orchestration}. This is executed through a container orchestrator enabling, e.g., deployments, scaling, auto-healing, and monitoring. A popular open-source container orchestrator is Kubernetes which provides software to build and deploy reliable, scalable distributed systems~\cite{burnsKubernetesRunningDive2019}.
Kubernetes works declaratively, meaning that instead of describing \textit{how} to perform operations, it is specified \textit{what} the desired state of the system should be. Kubernetes then takes care of the details required to achieve this state. If the actual state of the cluster deviates from the desired one, Kubernetes \textit{controllers} take action to reconcile the difference. \cite{burnsKubernetesRunningDive2019}, \cite{cncf_kubernetes}

\textit{Operators} in terms of Kubernetes' \textit{operator pattern} are software extensions to Kubernetes that let the developer extend the cluster's behavior without modifying the code of Kubernetes~\cite{kubernetes_operator}. A Kubernetes operator is a specific kind of a controller with domain-specific logic implemented inside~\cite{kopf_concepts}. It allows more sophisticated logic defining what should happen when a deviation between the desired and the actual state of the cluster is detected. Our concept of custom operators is based on the operator pattern. We implement the custom operators using the framework \textit{Kopf}~\cite{kopf_concepts}.

The need for orchestrating and managing services arises with vehicles becoming increasingly defined by software and with automotive service-oriented architectures (SOA) being developed~\cite{kampmann2022asoa}, \cite{kampmann_dynamic_2019}, \cite{rumezASOA2020}. Research is conducted with regard to in-vehicle orchestration~\cite{schindewolf_toward_2022}, \cite{nayak_automotive_2023} and also to orchestration on vehicle and infrastructure level where intelligence can be distributed in a cluster of computing units in vehicles, edge, cloud, and infrastructure~\cite{magaia_intelligent_2021}, \cite{arthurs_taxonomy_2022}. Our work contributes to the orchestration on \mbox{C-ITS} level enabling the management of complex applications spanning across multiple nodes and involving multiple time-shifted requests with a demand-driven approach.
In this paper, the term \textit{node} refers to the context of cloud computing~\cite{cncf_node} and describes a physical computer which may be employed in a C-ITS entity.

\begin{figure*}[t]
    \centering
    \includegraphics[width=0.85\linewidth]{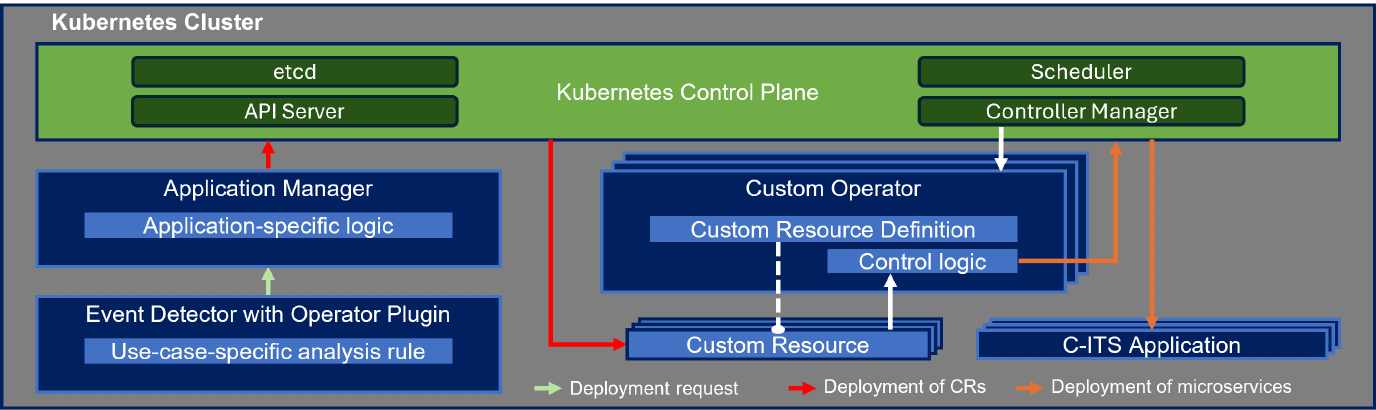}
    \caption{\textit{Reconciliation chain}: An event detector~\cite{reiherevent} detects events (patterns in data) according to user-defined analysis rules and sends a deployment request containing a description of the current demand from the \mbox{C-ITS}. The application manager interprets the request, adds further application-specific configuration, and deploys Kubernetes custom resources (CRs) encoding the current demand on the application via the Kubernetes API. The custom operators process the declared demand encoded in the CRs, consult their internal bookkeeping, and formulate a desired state of the application. By deploying, reconfiguring, or shutting down the application, action is taken to align the current state with the desired state.}
    \vspace{-3mm}
    \label{fig:concept_application_manager}
\end{figure*}

\section{Application Management in C-ITS}
\label{sec:application_management}
The demand-driven application management approach involves the \textit{application manager} and the \textit{custom operators} which work together to manage even complex applications.
The concept is designed based on requirements resulting from the context of \mbox{C-ITS}. Complex cooperative \mbox{C-ITS} applications may involve dynamic sets of multiple entities, requiring applications to be dynamically deployable, reconfigurable, and scalable. Efficient resource consumption is important, aiming to reduce energy, compute, and communication load.
Operating processes and applications on demand may reduce said loads, but requires coordination considering several past and present demands simultaneously.
Collective environment perception is an exemplary supporting offboard application requiring multiple \mbox{C-ITS} entities to share data with servers that process the data.
The supporting application may only be demanded by particular entities, e.g., vehicles, at certain times or regions of interest.
Also, since computing processes could be distributed across multiple computing nodes, this use case highlights the need for scalability while keeping applications and microservices node-agnostic.

The following key requirements for the developed approach were taken into account considering the management of complex cooperative applications in \mbox{C-ITS}:
\begin{itemize}
	\item \textit{Demand-driven orchestration}: Decide whether an application is to be deployed, reconfigured, updated, or scaled based on the current demands from the \mbox{C-ITS}.
	\item \textit{Bookkeeping capability}: Dynamically reconcile changing and new demands set by requesters. Account for time-shifted demands for the same application to avoid ambiguities or unnecessary replication.
	\item \textit{Environment-specific configuration}: Applications and their microservices are agnostic to their environment. Enable the (dynamic) (re)configuration of services allowing to respond to changing demands.
	\item \textit{Scalability}: The approach is capable of handling multiple nodes allowing distributed computing. It is agnostic to the number of involved \mbox{C-ITS} entities and services.
\end{itemize}

\subsection{Design Principles for Applications}
We propose a set of design principles for applications that our concept of application management is based on. The set is extensible and meant to be adapted for future use cases.

An \textit{application} is a set of one or multiple microservices while each microservice is packaged into a single container running independently from its environment. For example, an application for collective environment perception is composed of multiple microservices.
It might involve services for object detection, for fusing object lists, and for data transmission.
While an application can span across multiple nodes, data transmission between nodes is realized via \textit{connections} deployed as pairs of \textit{communication client services}. Connections are part of the application that they serve for and can be dynamically enabled, disabled, and configured.

Microservices offer \textit{interfaces} which allow for \textit{reconfiguration} also during runtime. One microservice can be part of several applications at once. For instance, a sensor driver service provides data for the two applications lane detection and object detection.

Applications are provided in the form of one or more \textit{container images} built through automated containerization processes~\cite{docker_ros}. A container image is application-agnostic and can be reused in different applications.
The Kubernetes resources necessary to run one or more containers in the cluster are packaged in \textit{Helm charts}~\cite{helm}. Helm charts offer parameterization allowing the deployment of containers to be adapted to the specific needs of the application.
Container images and Helm charts are provided in \textit{application registries} facilitating the integration of applications with continuous development processes. After development, an application has to pass certain steps of testing, including simulation~\cite{geller_carlos_2024}, before it is made available for deployment.

\subsection{Reconciliation Chain}
Our application management approach employs the \textit{reconciliation chain} extending the Kubernetes reconciliation loop where controllers constantly observe the current state of the cluster and take action to make the observed state match the desired state~\cite{burnsKubernetesRunningDive2019}.
Although controllers automate the reconciliation of current and desired states, the desired state of each Kubernetes resource must be defined individually, e.g., by the user.
The \textit{reconciliation chain} automates the declaration of desired states of individual Kubernetes resources and extends the reconciliation loop to the application level enabling the demand-driven deployment, reconfiguration, and scaling of complex multi-service \mbox{C-ITS} applications requested by multiple \mbox{C-ITS} entities.
The reconciliation chain, which is illustrated in Fig.~\ref{fig:concept_application_manager}, involves several components either developed by us or based on existing concepts of Kubernetes. The event detector with operator plugin~\cite{reiherevent}, the application manager, and custom operators are the key components.
Demands from the \mbox{C-ITS} are translated through the reconciliation chain into the desired state of an application, with subsequent actions taken to align the observed state with the desired state.

\subsection{Application Manager}
The application manager, together with an event detector with operator plugin~\cite{reiherevent}, forms an operator application~\cite{robotkube}. The event detector detects events, which are patterns in data exchanged in the \mbox{C-ITS}, according to implemented analysis rules. As a reaction upon an event, the event detector sends a \textit{deployment request} to the application manager. It contains a formulated \textit{demand} which may include information about \textit{which application} is demanded, basic application \textit{configurations}, necessary \textit{communication channels}, and \textit{which \mbox{C-ITS} entities request} the application.
The event detector sends this information without having any knowledge about previous states of (running) applications, including whether it is already running or what configuration it may already be running with. The decision whether the application's microservices are to be deployed, reconfigured, or shut down is made in the consecutive steps of the reconciliation chain.

The application manager interprets the deployment request and determines which specific microservices and communication channels are needed to enable the application to run. In this regard, further domain-specific configuration is added. Application-specific logic may be implemented to determine which microservices are distributed to specific computing nodes.
To sum up, the application manager condenses the information from the request into a declarative description of the current demand on the requested application. This information is then encoded in Kubernetes \textit{custom resources~(CRs)} which are deployed by the application manager via the Kubernetes API.

The deployment request is implemented as a ROS 2 action interface, with the application manager being the action server. With our extendable reference implementation, we publish source code for both action interface and server.

The application manager can act in certain \textit{access domains} on the nodes while not being allowed to interact with services outside the domain due to, e.g., safety or security reasons.

Furthermore, the application manager can orchestrate the rollout of updates and upgrades of applications on demand. Applications are updated, for example, by deploying new versions of the container images and/or Helm charts.

\subsection{Custom Operators}
The subsequent step of the reconciliation chain is executed by the custom operators which process the declared demands encoded in the CRs, consult their internal bookkeeping to take past states into account, and formulate a new desired state of the application.
Custom operators are implemented based on Kubernetes' operator pattern as described in Section~\ref{sec:related_work}.
In particular, an operator watches for deviations between the current state of a Kubernetes custom resource (CR) and its declared desired state. If a deviation occurs, some developer-defined logic, e.g., the deployment of an application, is executed by the operator~\cite{kubernetes_operator}.
The content, e.g., the configuration of the application to be deployed, that is contained in a CR of a specific kind, is defined through a \textit{custom resource definition (CRD)}~\cite{kubernetes_crd}. The developer specifically designs a CRD to contain distinct domain-specific information based on which the operator does its work.

Considering this customizability offered by Kubernetes, we apply the interaction of operator and CRD to manage the lifecycle of specific parts of an application. In our approach, CRDs are designed such that they contain certain information based on which the custom operator decides whether the concerned part of an application is to be deployed, reconfigured, or shut down. In particular, this information contains the requested configuration and the list of requesting \mbox{C-ITS} entities which both together form the demand on that specific part of an application at a certain point in time.
One CR, defined through a CRD, might represent a collection of tightly coupled services which are usually deployed together.

Thus, in terms of the reconciliation chain, if a demand on an application changes~--~e.g., modified configuration or set of requesters~--~the application manager deploys one or several new CRs or updates existing ones. The custom operators (one per CRD) observe the changes and take action to align the current state of the application with the desired state.

The custom operators implement a \textit{bookkeeping logic} with which they store the frequently received demands as data structs. This may contain which \mbox{C-ITS} entity requests which application and specific dynamic configurations. Based on the bookkeeping, it can be determined whether an application which might already be running is to be reconfigured instead of being deployed again.


\begin{figure*}[t]
  \centering
  \includegraphics[width=0.73\linewidth]{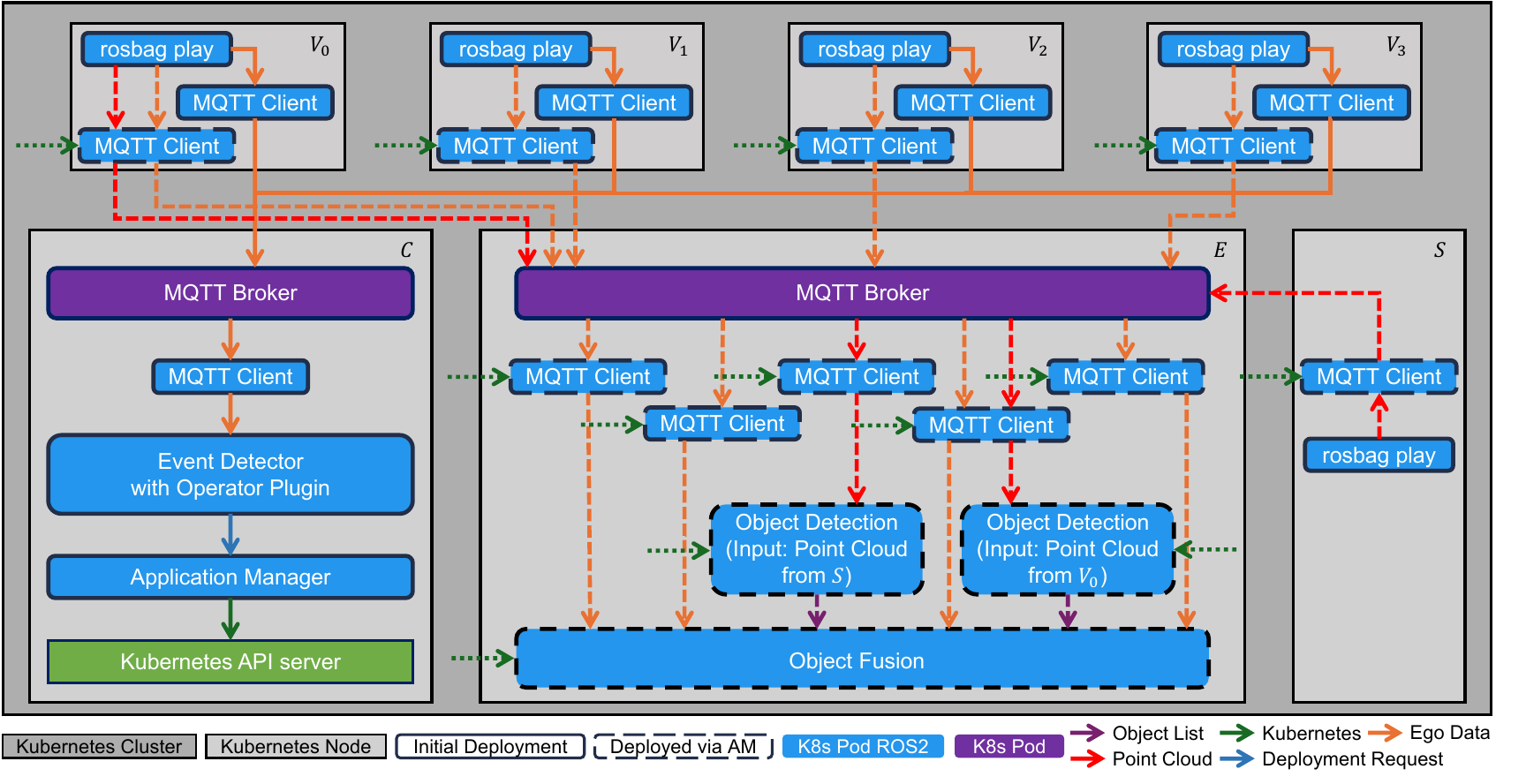}
  \vspace{-3mm}
  \caption{Experimental setup: $V_0$,\ldots, $V_3$ send ego data (orange) to $C$ via MQTT. An event detector analyzes the data and sends deployment requests (blue) to the application manager which triggers the Kubernetes API server. When vehicles approach the intersection, object detection and object fusion services are deployed on $E$. Point clouds (red dashed) and ego data (orange dashed) are transmitted from $S$ and $V_0$,\ldots, $V_3$ to $E$ on demand. Running on $E$, the object detection services receive point clouds and publish object lists (purple dashed). The object fusion service fuses object lists and ego data into one comprehensive object list which may be provided to the vehicles or other requesting \mbox{C-ITS} entities (not considered in our experiment).}
  \vspace{-4mm}
  \label{fig:experimental_setup}
\end{figure*}

\section{Experimental Setup}
To demonstrate its capabilities, we apply our approach in the scope of the \mbox{C-ITS} use case of collective environment perception involving a dynamically changing set of \mbox{C-ITS} entities.
We assume for our use case that the collectively perceived environment is represented in the form of an object list. To obtain that object list, we employ an application that runs both the detection of objects and the fusion of object lists and poses from multiple sources. In terms of distributed computing, services can run on various nodes. Thus, a service running data processing does not necessarily need to operate at the data source. The application involves multiple microservices (also visualized in Fig.~\ref{fig:experimental_setup}):
\begin{itemize}
  \item \textit{Object detection services} detecting objects based on raw sensor data which is in our experiment point clouds from lidar sensors. If $N$ is the number of sources providing point clouds, $N$ services are deployed on an edge server on demand if data sources are available.
  \item \textit{Object fusion service} fusing object lists and poses from multiple sources into one comprehensive object list. The set of subscribed input data topics is dynamically updated according to the available data sources. To achieve this, the service is reconfigured during runtime.
  \item \textit{Communication client services} being deployed pairwise on demand to establish communication channels (\textit{connections}) between nodes in the Kubernetes cluster transmitting certain data topics if required.
\end{itemize}

As listed in Table~\ref{tab:use_case_participants}, the experimental setup entails various \mbox{C-ITS} entities, such as connected vehicles (CVs), a sensor-equipped roadside infrastructure station unit (RISU), an edge server, and a cloud server. The CVs send information about their current state, \textit{ego data}~\cite{perception_interfaces}, including their poses. CV~$V_0$ and RISU $S$ are both equipped with lidar sensors and, therefore, provide point clouds. The setup is visualized in Fig.~\ref{fig:experimental_setup}. All entities are represented as nodes in a Kubernetes cluster. On the nodes, microservices run in Kubernetes pods.

\begin{table}[htbp]
  \caption{Participants of the collective Perception use case}
  \label{tab:use_case_participants}
  \centering
  \begin{tabularx}{\linewidth}{cXX}
    \toprule
     & \textbf{C-ITS Entity} & \textbf{Data Flow} \\
    \midrule
    $V_0$ & Connected vehicle (CV)                 & Send ego data, point cloud \\
    $V_1$ & Connected vehicle (CV)                 & Send ego data   \\
    $V_2$ & Connected vehicle (CV)                 & Send ego data   \\
    $V_3$ & Connected vehicle (CV)                 & Send ego data   \\
    $S$   & Roadside infrastructure station unit (RISU) & Send point cloud \\
    $E$   & Edge server                       & Receive ego data, point cloud \\
    $C$   & Cloud server              & Receive ego data \\
    \bottomrule
  \end{tabularx}
\end{table}

With the experiment, we aim to prove the capabilities of our presented approach taking the requirements listed in Section~\ref{sec:application_management} into account:
\begin{itemize}
	\item Demand-driven orchestration (including necessary connections between nodes)
	\item Bookkeeping capability
	\item Environment-specific configuration
	\item Scalability
\end{itemize}
The use case is constructed as follows:
\begin{itemize}
	\item $V_0$,\ldots, $V_3$ are part of the \mbox{C-ITS} and follow their routes.
	\item $S$ is located at a challenging intersection. Collective perception is applied to enhance situational awareness for \mbox{C-ITS} entities nearby if at least one recipient (CV in our use case) approaches the intersection.
	\item $E$ is capable of running computationally expensive tasks. The services object detection and object fusion are deployed on $E$ on demand. Data transmission from data sources to $E$ is enabled on demand.
	\item Running on $C$, the event detector (ED) continually analyzes the CVs' poses. Once at least one CV is within a defined distance $d_\text{start}$ to the intersection, a deployment request for the \textit{object detection fusion application} is sent to the application manager. The services are deployed on the CVs, on $S$ and on $E$.
	\item With further CVs approaching the intersection, the ED sends additional requests for the \textit{object detection fusion application}. The application manager and the custom operators decide whether services need to be deployed or reconfigured. Their bookkeeping ensures that services, even though requested multiple times, are deployed only once. The set of input topics for the object fusion service is dynamically reconfigured.
	\item When a CV (requester of the application) leaves the intersection, the ED sends a request to shut down the application. The CV is removed from the bookkeeping ensuring only those services which are no longer demanded by any requester are shut down.
	\item When the last CV has left the intersection, the bookkeeping is empty and no requesters are left. The application with its services is shut down.
\end{itemize}

The Kubernetes cluster is set up using \textit{k3d}~\cite{k3d} which allows multi-node clusters while running on a single machine. The software components are based  on ROS 2. ROS 2 bags recorded in simulation are used to simulate live data from the CVs and the RISU. \textit{MQTT} is used for data transmission across nodes. To allow reproducibility, the experimental setup is made publicly available \footnote{\url{https://github.com/ika-rwth-aachen/robotkube/} \newline(use case ``Collective Perception at Intersection'')}.

\begin{table*}[!htbp]
    \caption{Evaluation of the bookkeeping resulting from demands examined in the collective perception use case (Connections represent the communication channels between two nodes, with the direction of the data flow indicated in brackets)}
    \label{tab:use_case}
    \centering
    \scriptsize
    \begin{tabularx}{\textwidth}{p{0.3cm}p{1.5cm}p{6.1cm}p{4.8cm}p{3.3cm}}
      \toprule
      \textbf{Step} & \textbf{Event} & \textbf{Deployment Request: Current Demand} & \textbf{Resulting State of Bookkeeping: Requesters} & \textbf{Scenario Illustration} \\
      \midrule
      1 & \raggedright $V_0$ approaches $S$ at intersection & \textbf{Application}: \textit{Object detection fusion} with the following input topics: Ego data ($V_0$), point cloud ($V_0$ and $S$) \newline \textbf{Connection} \texttt{($V_0 \rightarrow E$)}: Ego data, point cloud \newline \textbf{Connection} \texttt{($S \rightarrow E$)}: Point cloud & \textbf{Object Detection 1} \texttt{($S$)}: $V_0$, $S$ \newline \textbf{Object Detection 2} \texttt{($V_0$)}: $V_0$, $S$ \newline \textbf{Object Fusion}: $V_0$, $S$ \newline \textbf{Connection 1} \texttt{($V_0 \rightarrow E$)}: $V_0$, $S$ \newline \textbf{Connection 2} \texttt{($S \rightarrow E$)}: $V_0$, $S$ &
      \raisebox{-14.5mm}[3mm]{\includegraphics[width=0.85\linewidth]{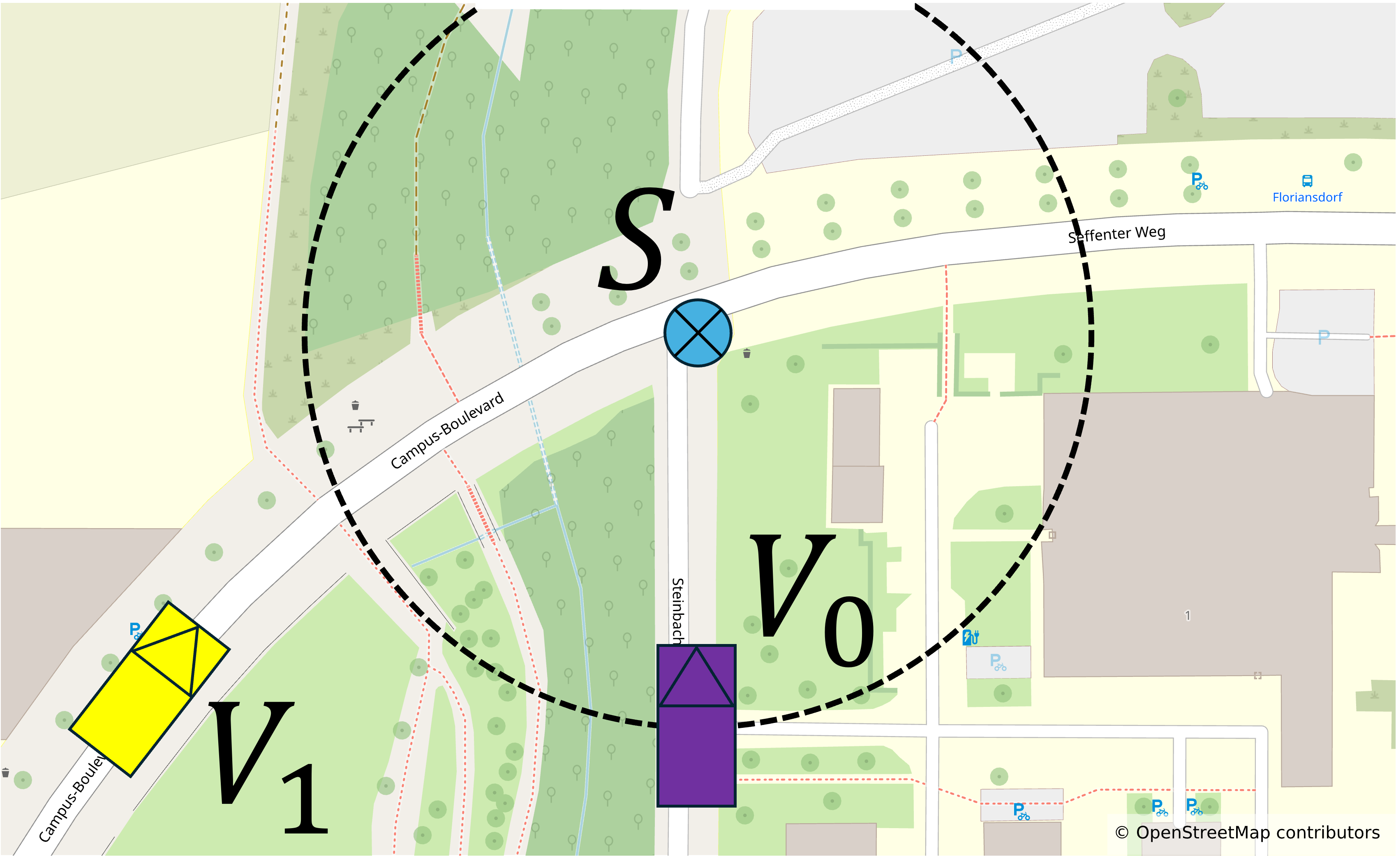}}
      \\
      \hline
      2 & \raggedright $V_1$ approaches $S$ at intersection & \textbf{Application}: \textit{Object detection fusion} with the following input topics: Ego data ($V_1$), point cloud ($S$) \newline \textbf{Connection} \texttt{($V_1 \rightarrow E$)}: Ego data \newline \textbf{Connection} \texttt{($S \rightarrow E$)}: Point cloud & \textbf{Object Detection 1} \texttt{($S$)}: $V_0$, $S$, $V_1$ \newline \textbf{Object Detection 2} \texttt{($V_0$)}: $V_0$, $S$ \newline \textbf{Object Fusion}: $V_0$, $S$, $V_1$ \newline \textbf{Connection 1} \texttt{($V_0 \rightarrow E$)}: $V_0$, $S$ \newline \textbf{Connection 2} \texttt{($S \rightarrow E$)}: $V_0$, $S$, $V_1$ \newline \textbf{Connection 3} \texttt{($V_1 \rightarrow E$)}: $V_1$, $S$ &
      \raisebox{-14.5mm}[3mm]{\includegraphics[width=0.85\linewidth]{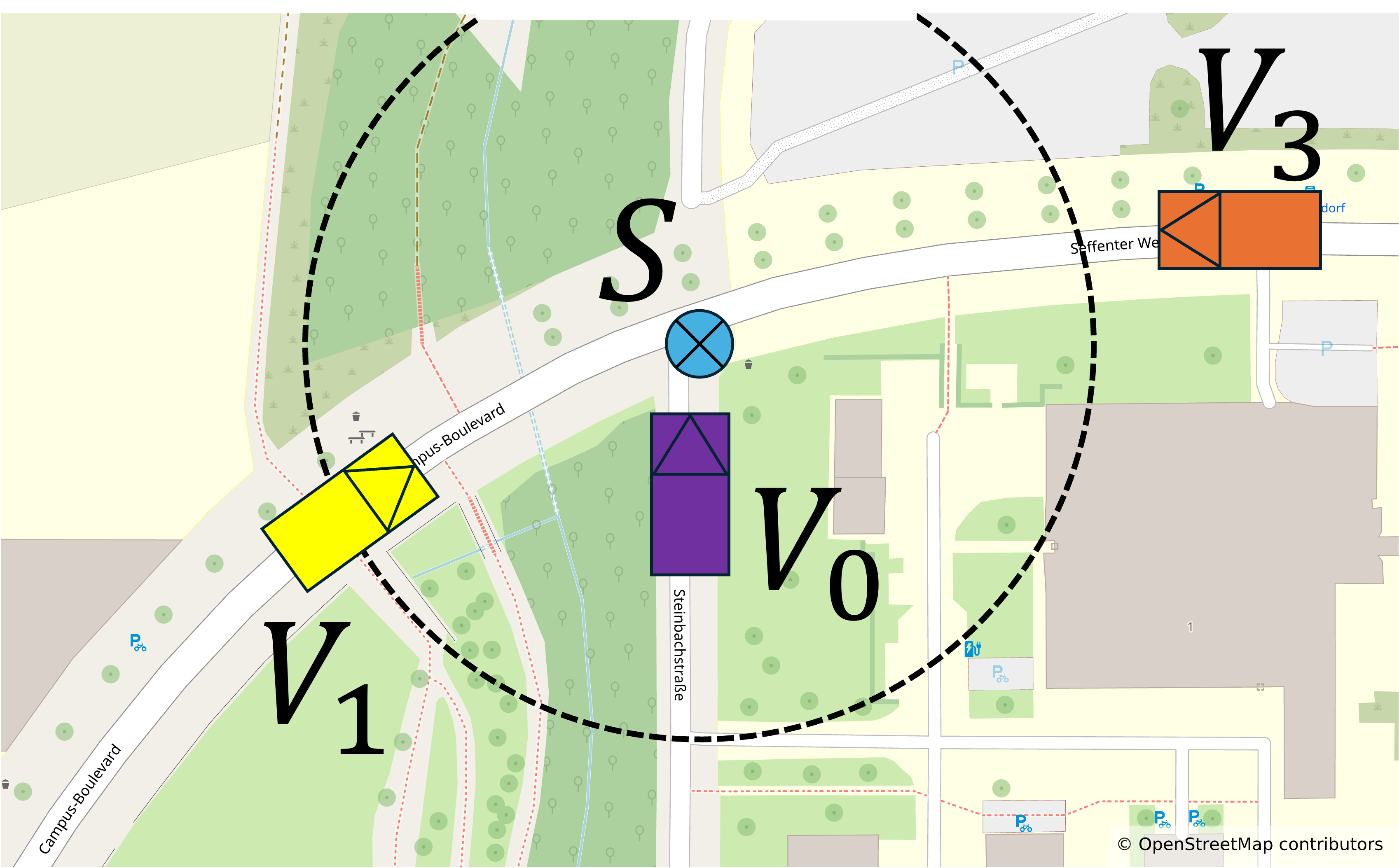}}
      \\
      \hline
      3 & \raggedright $V_3$ approaches $S$ at intersection & \textbf{Application}: \textit{Object detection fusion} with the following input topics: Ego data ($V_3$), point cloud ($S$) \newline \textbf{Connection} \texttt{($V_3 \rightarrow E$)}: Ego data \newline \textbf{Connection} \texttt{($S \rightarrow E$)}: Point cloud & \textbf{Object Detection 1} \texttt{($S$)}: $V_0$, $S$, $V_1$, $V_3$ \newline \textbf{Object Detection 2} \texttt{($V_0$)}: $V_0$, $S$ \newline \textbf{Object Fusion}: $V_0$, $S$, $V_1$, $V_3$ \newline \textbf{Connection 1} \texttt{($V_0 \rightarrow E$)}: $V_0$, $S$ \newline \textbf{Connection 2} \texttt{($S \rightarrow E$)}: $V_0$, $S$, $V_1$, $V_3$ \newline \textbf{Connection 3} \texttt{($V_1 \rightarrow E$)}: $V_1$, $S$ \newline \textbf{Connection 4} \texttt{($V_3 \rightarrow E$)}: $V_3$, $S$ &
      \raisebox{-14.5mm}[3mm]{\includegraphics[width=0.85\linewidth]{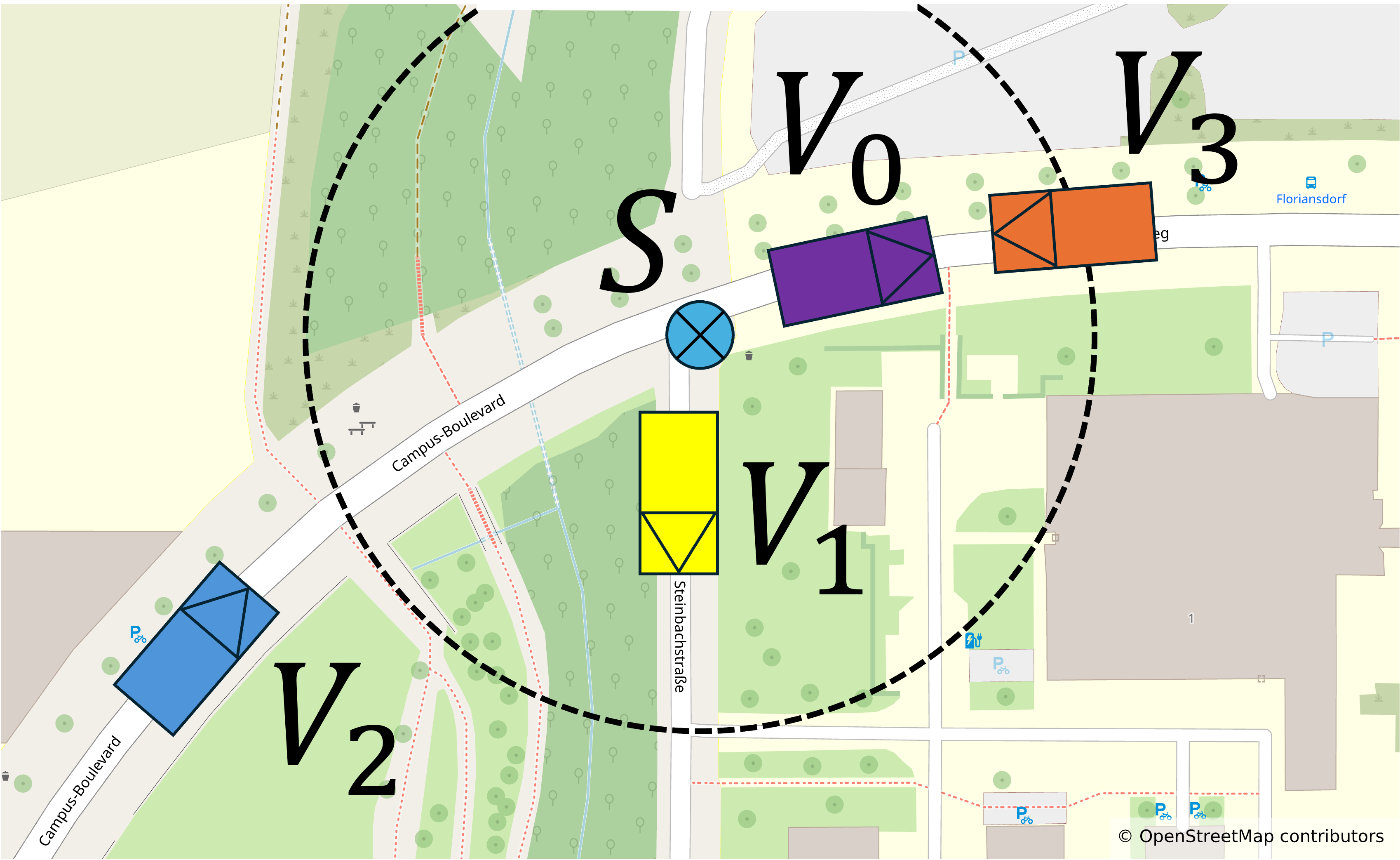}}
      \\
      \hline
      4 & \raggedright $V_2$ approaches $S$ at intersection & \textbf{Application}: \textit{Object detection fusion} with the following input topics: Ego data ($V_2$), point cloud ($S$) \newline \textbf{Connection} \texttt{($V_2 \rightarrow E$)}: Ego data \newline \textbf{Connection} \texttt{($S \rightarrow E$)}: Point cloud & \textbf{Object Detection 1} \texttt{($S$)}: $V_0$, $S$, $V_1$, $V_2$, $V_3$ \newline \textbf{Object Detection 2} \texttt{($V_0$)}: $V_0$, $S$ \newline \textbf{Object Fusion}: $V_0$, $S$, $V_1$, $V_2$, $V_3$ \newline \textbf{Connection 1} \texttt{($V_0 \rightarrow E$)}: $V_0$, $S$ \newline \textbf{Connection 2} \texttt{($S \rightarrow E$)}: $V_0$, $S$, $V_1$, $V_2$, $V_3$ \newline \textbf{Connection 3} \texttt{($V_1 \rightarrow E$)}: $V_1$, $S$ \newline \textbf{Connection 4} \texttt{($V_3 \rightarrow E$)}: $V_3$, $S$ \newline \textbf{Connection 5} \texttt{($V_2 \rightarrow E$)}: $V_2$, $S$ &
      \raisebox{-14.5mm}[3mm]{\includegraphics[width=0.85\linewidth]{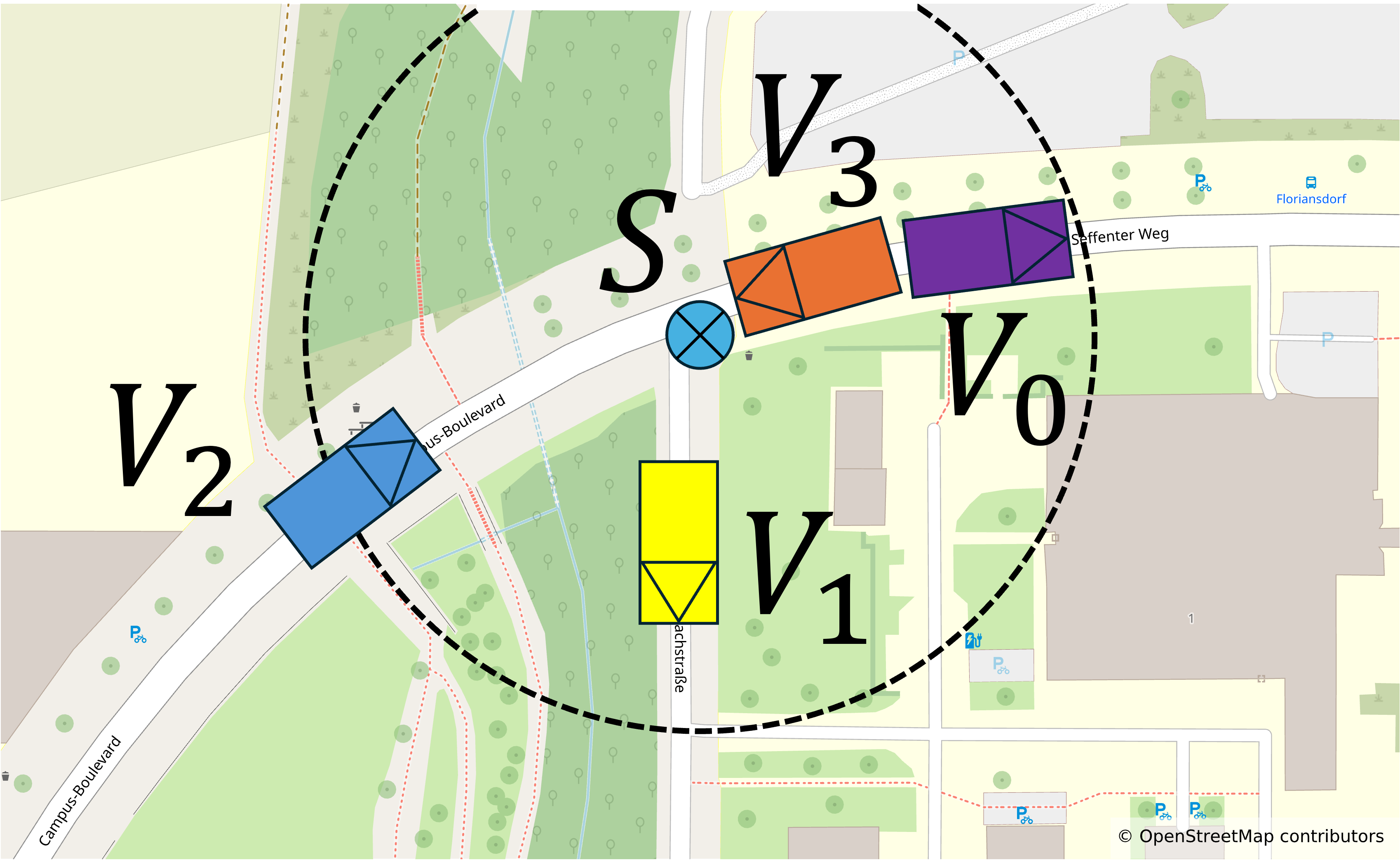}}
      \\
      \hline
      5 & \raggedright $V_0$ leaves intersection & \textbf{Application [Shutdown]}: \textit{Object detection fusion} with the following input topics: Ego data ($V_0$), point cloud ($V_0$ and $S$) \newline \textbf{Connection [Shutdown]} \texttt{($V_0 \rightarrow E$)}: Ego data, point cloud \newline \textbf{Connection [Shutdown]} \texttt{($S \rightarrow E$)}: Point cloud & \textbf{Object Detection 1} \texttt{($S$)}: $S$, $V_1$, $V_2$, $V_3$ \newline \textbf{Object Fusion}: $S$, $V_1$, $V_2$, $V_3$ \newline \textbf{Connection 2} \texttt{($S \rightarrow E$)}: $S$, $V_1$, $V_2$, $V_3$ \newline \textbf{Connection 3} \texttt{($V_1 \rightarrow E$)}: $V_1$, $S$ \newline \textbf{Connection 4} \texttt{($V_3 \rightarrow E$)}: $V_3$, $S$ \newline \textbf{Connection 5} \texttt{($V_2 \rightarrow E$)}: $V_2$, $S$ &
      \raisebox{-14.5mm}[3mm]{\includegraphics[width=0.85\linewidth]{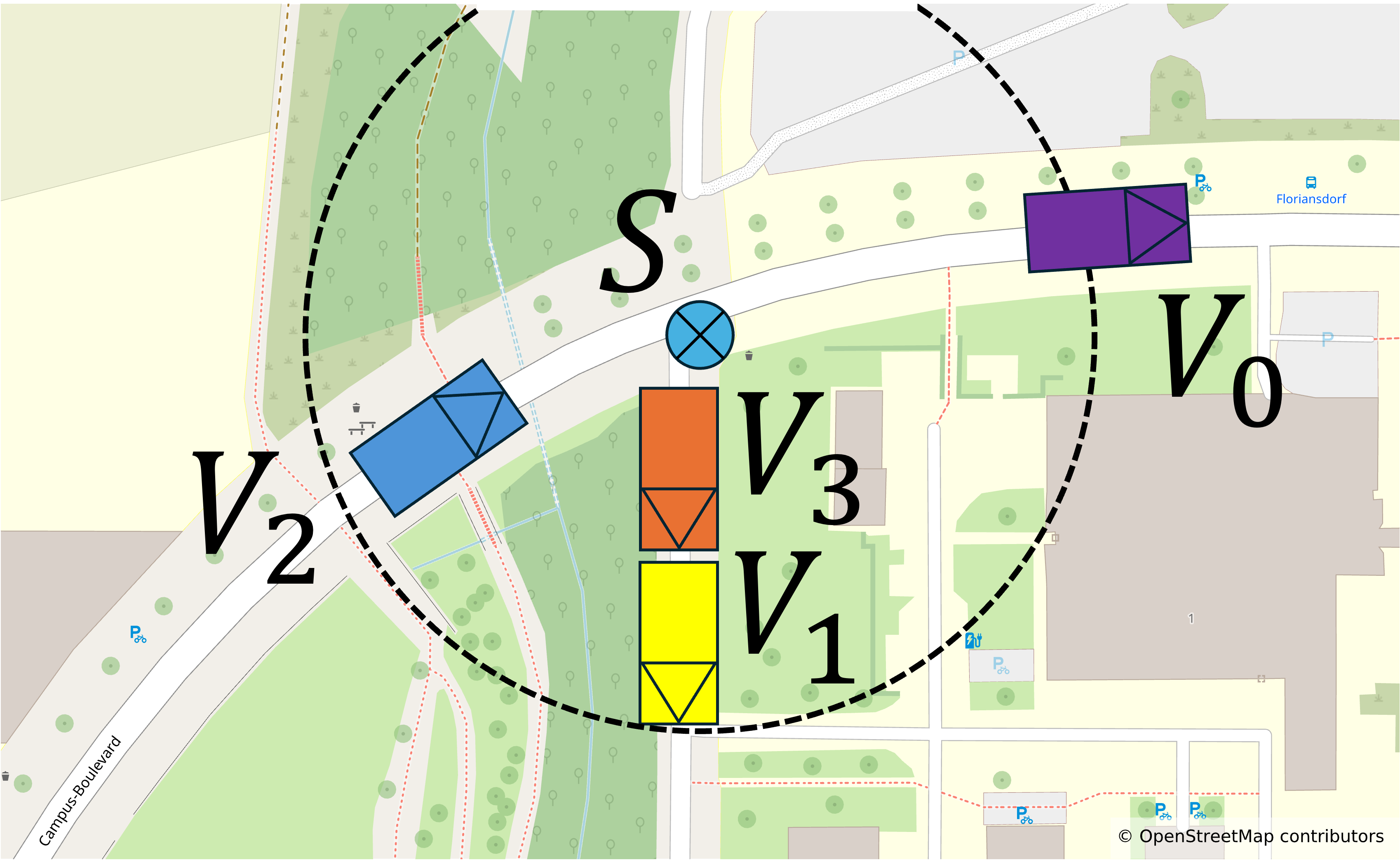}}
      \\
      \hline
      6 & \raggedright $V_1$ leaves intersection & \textbf{Application [Shutdown]}: \textit{Object detection fusion} with the following input topics: Ego data ($V_1$), point cloud ($S$) \newline \textbf{Connection [Shutdown]} \texttt{($V_1 \rightarrow E$)}: Ego data \newline \textbf{Connection [Shutdown]} \texttt{($S \rightarrow E$)}: Point cloud & \textbf{Object Detection 1} \texttt{($S$)}: $S$, $V_2$, $V_3$ \newline \textbf{Object Fusion}: $S$, $V_2$, $V_3$ \newline \textbf{Connection 2} \texttt{($S \rightarrow E$)}: $S$, $V_2$, $V_3$ \newline \textbf{Connection 4} \texttt{($V_3 \rightarrow E$)}: $V_3$, $S$ \newline \textbf{Connection 5} \texttt{($V_2 \rightarrow E$)}: $V_2$, $S$ &
      \raisebox{-14.5mm}[3mm]{\includegraphics[width=0.85\linewidth]{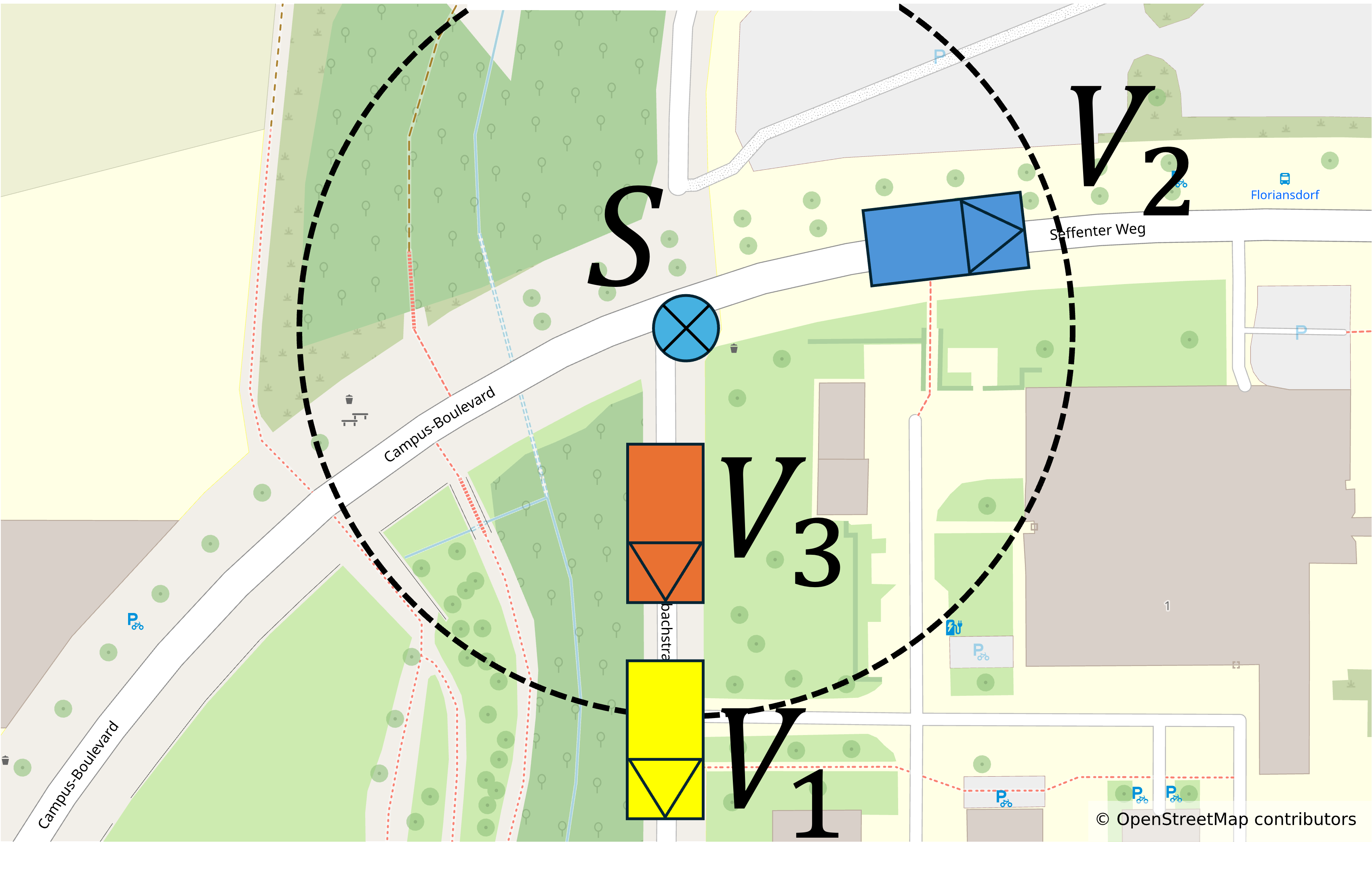}}
      \\
      \hline
      7 & \raggedright $V_2$ leaves intersection & \textbf{Application [Shutdown]}: \textit{Object detection fusion} with the following input topics: Ego data ($V_2$), point cloud ($S$) \newline \textbf{Connection [Shutdown]} \texttt{($V_2 \rightarrow E$)}: Ego data \newline \textbf{Connection [Shutdown]} \texttt{($S \rightarrow E$)}: Point cloud & \textbf{Object Detection 1} \texttt{($S$)}: $S$, $V_3$ \newline \textbf{Object Fusion}: $S$, $V_3$ \newline \textbf{Connection 2} \texttt{($S \rightarrow E$)}: $S$, $V_3$ \newline \textbf{Connection 4} \texttt{($V_3 \rightarrow E$)}: $V_3$, $S$ &
      \raisebox{-14.5mm}[3mm]{\includegraphics[width=0.85\linewidth]{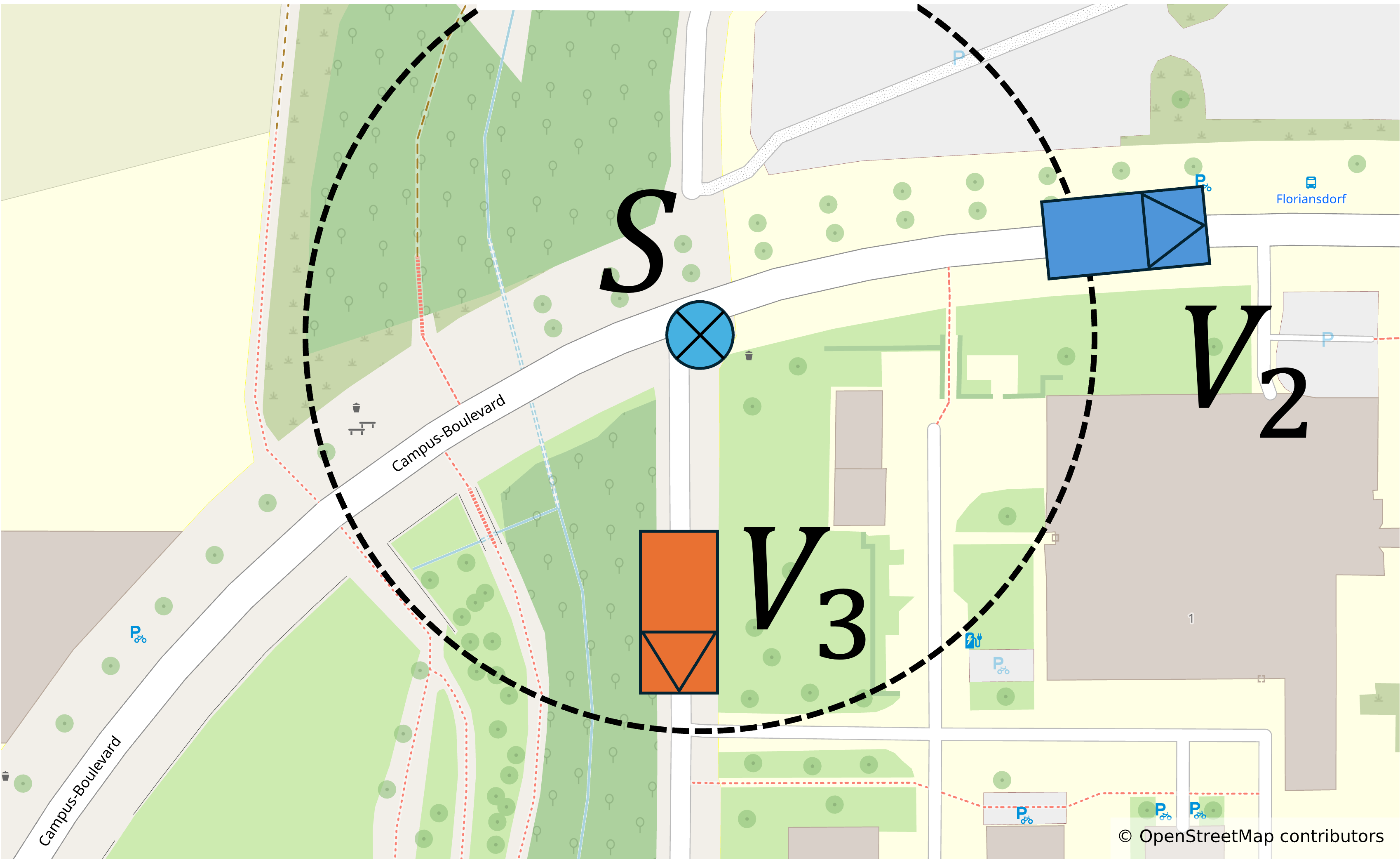}}
      \\
      \hline
      8 & \raggedright $V_3$ leaves intersection & \textbf{Application [Shutdown]}: \textit{Object detection fusion} with the following input topics: Ego data ($V_3$), point cloud ($S$) \newline \textbf{Connection [Shutdown]} \texttt{($V_3 \rightarrow E$)}: Ego data \newline \textbf{Connection [Shutdown]} \texttt{($S \rightarrow E$)}: Point cloud &  &
      \raisebox{-14.5mm}[3mm]{\includegraphics[width=0.85\linewidth]{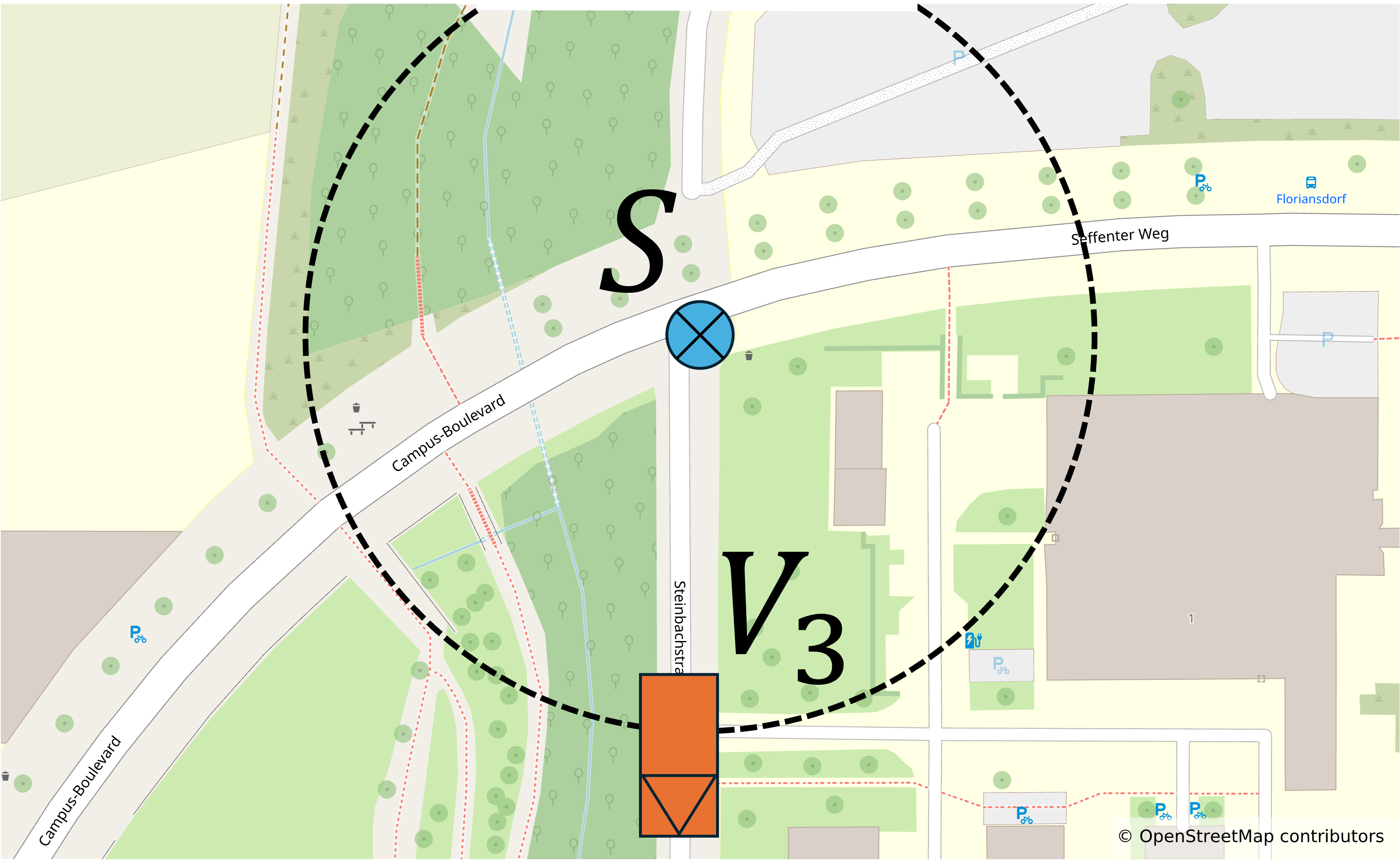}}
      \\
      \bottomrule
    \end{tabularx}
\end{table*}

\section{Evaluation}
The experimental setup is used to evaluate the capabilities of our approach for application management orchestrating the demand-driven deployment and reconfiguration of the \textit{object detection fusion application} in a dynamic environment.

For evaluation, while running the experiment with the CVs approaching and leaving the intersection, the requesters of the services stored in the bookkeeping are monitored. Also, the currently demanded configurations of the services, e.g., the requested input topics for the object fusion service, are monitored.
To verify whether the connections are established correctly, the incoming topics on $E$ are observed.
Furthermore, it is examined whether the services are deployed on the correct Kubernetes nodes. Table~\ref{tab:use_case} lists the steps of our constructed use case containing the events, the resulting deployment requests with the current demands for the application, and the resulting state of the bookkeeping. The outcomes of the conducted experiment are discussed with respect to the formulated requirements.

\textit{Demand-driven orchestration}: The event detector (ED) is able to detect the vehicles approaching the intersection. As outlined in the third column in Table~\ref{tab:use_case}, for each step, the ED formulates a deployment request containing the current demand on the application resulting from the event. ``Current demand'' means that the ED does not have any knowledge about the previous deployments. The ED does not consider whether the application is already running.
This is done by the application manager (AM). The comparison of the fourth with the third column shows that the AM correctly decides which concrete services need to be either deployed or reconfigured. As a result, the sequence of the deployment of the services corresponds to the sequence of the events.

\textit{Bookkeeping capability}: The list of requesters contained in the bookkeeping for each service is shown in the fourth column in Table~\ref{tab:use_case}. The ability of the bookkeeping to track current demands and reconcile them with past demands can be verified with the list of requesters per service. E.g., in ``Step 2'', with the arrival of $V_1$, object detection on point cloud from $S$ is demanded (see ``Object Detection 1''). Since this service has already been deployed, instead of deploying it again, $V_1$ is added to the list of requesters for this service. The same applies for the connections.

\textit{Environment-specific configuration}: The object fusion service, which needs to be reconfigured dynamically during runtime to adapt to a dynamically changing set of input topics, is investigated. The set of input topics changes with CVs approaching or leaving the intersection.
By inspecting the ROS 2 node information, including the list of subscribed topics, it is examined that the object fusion service is reconfigured with the correct input topics according to the current list of requesters. Its configuration is updated dynamically during runtime without restarting the service.
Furthermore, it is observed that all expected data topics are received on $E$ underlining that the configuration of the connections is done correctly.
Said aspects provide evidence that, assumed a service offers proper interfaces, dynamic and environment-specific (re)configuration is achieved with our approach.

\textit{Scalability}: Scalability is demonstrated with the design of the experiment involving multiple \mbox{C-ITS} entities acting as sources for data and demand.
The experiment shows that the demand-driven orchestration works in principle independently from the number of CVs or data sources. Furthermore, complexity was added by involving an edge server in the setup highlighting that our approach enables distributed computing across multiple nodes.
Note that our framework is not limited to the use case of collective environment perception but it is designed to be extensible for arbitrary applications with various entities, data flows, and services. 


\section{Conclusion}
The presented approach for demand-driven application management in \mbox{C-ITS} allows to deploy, dynamically reconfigure, scale, update and upgrade applications based on demands originating from multiple entities in a \mbox{C-ITS}.

Our application management framework~--~comprising the \textit{application manager} and the \textit{custom operators}~--~implements a \textit{reconciliation chain}. This is a concept inspired by Kubernetes' reconciliation loop which is responsible for aligning the observed state of a cluster with a declared desired state. The reconciliation chain extends the reconciliation loop to the application level and is responsible for translating demands from the \mbox{C-ITS} into the desired state of an application involving multiple microservices.
Changing or new demands for an (already running) application are dynamically reconciled through our framework.
For example, time-shifted demands for the same application are handled and, therefore, unnecessary replication is avoided.

The approach is successfully applied in the complex cooperative \mbox{C-ITS} use case of collective environment perception. The key capabilities of the approach are demonstrated as a result of the conducted experiment: \textit{demand-driven orchestration}, \textit{bookkeeping capability}, \textit{environment-specific configuration}, and \textit{scalability}.

To allow reproducibility, we publish the source code of both our prototypical implementation of an application management framework and the experimental setup.

\section*{ACKNOWLEDGMENT}

This research was conducted within the research projects ``autotech.agil'' (FKZ~1IS22088A) and ``6GEM'' (FKZ~16KISK036K), funded by BMFTR (Federal Ministry of Research, Technology and Space), and is continued in the project ``iEXODDUS'' (EU Horizon, GA No.~101146091).


\bibliographystyle{IEEEtran}
\bibliography{root}

\end{document}